# PreNeT: Leveraging Computational Features to Predict Deep Neural Network Training Time

Alireza Pourali, Arian Boukani, Hamzeh Khazaei
York University
{alirezap,arbo,hkh}@yorku.ca

## ABSTRACT

Training deep learning models, particularly Transformer-based architectures such as Large Language Models (LLMs), demands substantial computational resources and extended training periods. While optimal configuration and infrastructure selection can significantly reduce associated costs, this optimization requires preliminary analysis tools. This paper introduces PreNeT, a novel predictive framework designed to address this optimization challenge. PreNeT facilitates training optimization by integrating comprehensive computational metrics, including layer-specific parameters, arithmetic operations and memory utilization. A key feature of PreNeT is its capacity to accurately predict training duration on previously unexamined hardware infrastructures, including novel accelerator architectures. This framework employs a sophisticated approach to capture and analyze the distinct characteristics of various neural network layers, thereby enhancing existing prediction methodologies. Through proactive implementation of PreNeT, researchers and practitioners can determine optimal configurations, parameter settings, and hardware specifications to maximize cost-efficiency and minimize training duration. Experimental results demonstrate that PreNeT achieves up to 72% improvement in prediction accuracy compared to contemporary state-of-the-art frameworks.

## 1 INTRODUCTION

Recent advances in machine learning have led to the creation of sophisticated models, including neural networks and Large Language Models (LLMs), that contain billions of parameters [5, 8, 33, 40, 41]. Progress has led advancements in natural language processing and computer vision, though, with increased resource requirements and prolonged training times [21, 31, 34]. This presents obstacles for researchers and professionals in relation to resource management, cost control, and proper project planning [37].

Like ordinary software systems, deep learning models come with diverse settings that require proper configuration. Practitioners often have to experiment with numerous settings like learning rate, batch sizes, and architectural components to find the best setup for a specific task (usually by using AutoML tools). Finding the optimal set of hyperparameters and architectures can be costly in practice, particularly with deeper networks like LLMs due to their high model complexity [9].

The field of AI accelerators has seen advancements in various hardware choices designed specifically for deep learning tasks emerging quickly over time. Besides GPUs that come in a range of sizes, from desktop versions to high-performance server models, specialized accelerators such as Google's TPUs [17] and Intel's Habana Labs Gaudi and Gaudi2 [14], Graphcore IPUs [12] and Cerebras Wafer Scale Engine (WSE) [3] have become notable alternatives while offering substantial power. These specialized processors are designed to meet the increasing requirements for training complex deep neural networks (DNN). With a variety of hardware choices for researchers and professionals alike comes the challenge of deciding that *which configuration is the optimal choice that delivers the best mix of performance efficiency and cost-effectiveness for training a DNN model?*

In recent years, there has been extensive research focused on predicting the runtime performance of machine learning models [10, 32, 45–47]. Our study aims to fill a gap in machine learning by delving into the prediction of training time for deep learning models across various complexities, from fundamental structures to more complex models like Transformers and Large Language Models (LLMs). While previous studies have focused on some aspect of performance prediction, such as calculations using units like single precision floating point operations (FLOPs) [17], they often fall short in capturing the complexity involved in training contemporary architectures. In our work, we use various machine learning techniques to assess and forecast the training time of the key components of deep learning models, such as Multi-Head Self-Attention, Embedding, and Normalization layers. PreNeT includes computational features like arithmetic operations, memory usage and benchmark input data to provide a richer understanding of the training process. To the best of our knowledge, this is the first method to consider the computational and layerwise complexity of machine learning models, including the recent complex models such as LLMs for generating training time predictions.

Specifically, in this paper, we investigate the following research questions:

**RQ1:** How do layer-specific operations such as Multi-Head Self-Attention affect the training time prediction of larger scale models such as LLMs? In large-scale models, layer-specific operations contribute heavily to computational load. Components like Multi-Head Self-Attention are particularly resource-intensive due to both the arithmetic operations and memory overhead involved. By incorporating these operations into our framework, we take account of the complexities of these models, which results in more accurate estimates, both in general and specifically tailored to suit large-scale applications like transformer-based models such as LLMs.

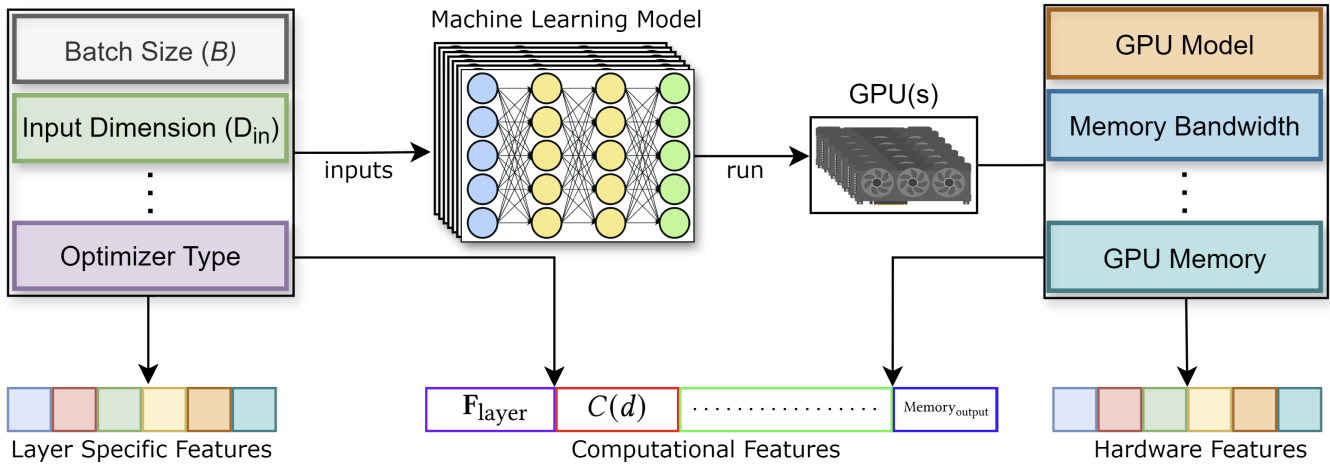

Figure 1: PreNeT Framework Overview

**RQ2:** How does integrating computational features such as arithmetic operations, memory usage, and layer-specific inputs enhance training time prediction for deep learning models? PreNeT is equipped with a wide range of computational data, including memory usage (the memory required when a model is running), arithmetic operations (catching the computational complexity of each layer), and layer-specific inputs to provide insights into the characteristics and size of various type of layers to enhance the training time predictions. By incorporating these features, PreNeT offers a much more detailed understanding of the elements and how they influence the training time. Hence, it can accurately predict across diverse architectures, such as traditional neural networks to more complex architectures like Transformers that come with a wider range of parameters during their execution stages.

**RQ3:** How can the computational features and layer-specific operations be leveraged to create a prediction model that estimates training times for unseen model configurations and hardware setups? We use data from various types of GPUs, including local RTX hardware and Google Cloud infrastructure, to create a reliable prediction model. By including hardware setups and model designs, we develop a system that can predict training times for new configurations or models, making it versatile and useful for real-world scenarios. Our method works well with unseen model configurations and hardware setups from simple neural networks to complex LLMs as illustrated in 5.2.

## 2 PROBLEM STATEMENT

Predicting the training performance of ML models is a complicated task due to the various model architectures, input hyper-parameters and specific hardware configurations used. When the complexity of the model is high (such as transformer-based models like LLMs), traditional methods based on a single metric FLOP often fail [35]. Contemporary methods take memory bandwidth into account, ensuring that data loading is factored in to provide more accurate performance estimations. Accurately predicting training performance is crucial for efficiently allocating resources, avoiding costly errors, and optimizing the training of the models across various hardware configurations.

Recent advances in large language models have led to the development of more complex architectures. Take the GPT 3 model [2] as an example with its impressive 175 billion parameter count. It emphasizes this shift by highlighting how deep learning can produce text that closely resembles expression. Training models like GPT 3 require a lot of computing power, and it could take up to weeks to train them on a massive GPU farm. This shows the importance of managing resources in deep learning tasks. Additionally, fine-tuned models that are based on user feedback are becoming more popular, which adds an extra layer to the complexity of the training process [30].

The primary goal of this work is to answer the question of predicting **how do different model architectures, hyper-parameters, and hardware configurations affect the training time?** These include considerations of architectural complexity, including a large number of Multi-Head Self-Attention and Normalization layers, hardware flexibility with single GPU machines for local work through to cloud-scale systems and hyper-parameter selections such as batch size, optimizer choice and learning rates. These factors significantly impact the training time, making accurate predictions essential for efficient resource allocation and cost savings.

***Which GPU to rent or purchase?*** In industry and academia, users could access various accelerators to run their experiments. Cloud providers offer a variety of GPUs for rent, each at different price points. The primary question remains: Which GPU should be used for the experiments? Which GPU could be the cheapest and fastest option for running the experiments? Which configuration is the best for a certain architecture? Using PreNeT, a user can easily feed the input data for a certain architecture and get a prediction of the training time per epoch to make better choices. By using this data and calculating the fees per GPU, users can make more informed choices when running their training jobs. The results in 5.2 clarify that our approach will provide better insights into the proper configuration and parameter settings for the training job.

Table 1: Overview of Features in PreNeT for Training Time Prediction by Layer Type

| Feature | Layer Type | Description |
|---|---|---|
| **Hardware Features** | | |
| GPU Model | All Layers | One-hot encoding of GPU architecture |
| Memory Bandwidth (GB/s) | All Layers | GPU memory bandwidth |
| Clock Speed (MHz) | All Layers | GPU clock speed |
| CUDA Cores | All Layers | Number of CUDA cores |
| GPU Memory (GB) | All Layers | Available memory on the GPU |
| **Layer-Specific Features** | | |
| Batch Size ($B$) | All Layers | Number of samples per batch |
| Input Dimension ($D_{in}$) | Dense, Convolution, Attention, Embedding, RNN | Size of input tensor |
| Output Dimension ($D_{out}$) | Dense, Convolution, Attention, Embedding | Size of output tensor |
| Kernel Size ($K$) | Convolutional | Size of the convolution filter |
| Sequence Length ($L$) | Attention, RNN, Transformer | Length of input sequence |
| Number of Attention Heads ($H$) | Attention, Transformer | Number of Attention heads in Multi-Head Attention |
| Embedding Dimension ($D_{emb}$) | Attention, Embedding | Dimensionality of embeddings or Attention heads |
| Bidirectionality | RNN | The direction of RNN input processing |
| CO | All Layers | Computational operations |
| **Memory Features** | | |
| Memory Weights ($D_{in} \times D_{out}$) | Dense, Convolution, RNN | Memory required for layer weights |
| Memory Input ($B \times D_{in}$) | Dense, Convolution, Attention, Embedding, RNN | Memory required for the input tensor |
| Memory Output ($B \times D_{out}$) | Dense, Convolution, Attention, Embedding | Memory required for the output tensor |
| **Training Configuration Features** | | |
| Dropout Rate | Attention, Transformer, RNN | Dropout rate applied during training |
| Learning Rate | All Layers | Learning rate used during training |
| Optimizer Type | All Layers | Type of optimizer used (e.g., Adam, SGD) |

## 3 METHODOLOGY

Here, we outline how to predict the training time for traditional machine learning models and deep learning models, including transformer-based models (e.g., LLMs), by integrating properties which reflect the computational complexity, hardware configurations and other key parameters that affect training time. As illustrated in Table 1, our method uses various features such as the architectural elements, computational complexity for each specific layer and GPU specifications to make predictions. Since the nature of these features is non-linear as specified in [18], we feed this data into various machine learning models and choose the best performing one for making the predictions. As shown in Figure 1, PreNeT is built upon Layer-Specific (3.1), Computational Operations (3.2) and Computational Memory (3.3) features.

### 3.1 LAYER-SPECIFIC FEATURES

During the benchmarking phase, to predict the training time for each architecture, we trained each layer of a deep learning model, including transformer-based models, on various GPUs. We systematically measured the runtime of many different neural network layers using our local and Google Cloud infrastructure. Every single layer (Attention, Embedding, Convolution, Dense, Normalization and RNN-based layers) was evaluated in terms of its running performance across seven different GPU architectures. The benchmarking process involved testing various parameters, such as batch size, input size, and sequence length, across each layer. We performed each run multiple times with 20,000 randomly generated iterations of the mixture of parameters and took the median runtime to ensure that the measurements were reliable and consistent. For each GPU and layer type, we generated 20,000 random configurations without any dependencies on previously selected features. This process resulted in a total of 140,000 different input parameters, for which we recorded the training times for all GPUs by layer type.

Our benchmarking data is thorough and examines the performance of various layer types across multiple GPU architectures. This detailed information allows us to understand the specific computational requirements unique to each layer type and GPU combination. The input parameters for each layer type lead to different training times on each GPU. By analyzing this data, we can consider the complexity of each layer type across different GPUs using a more comprehensive strategy. Inspired by [18], we used the layer-specific collected data to construct a PreNeT modelling capable of predicting training times per epoch. By taking these unique features for each layer type, we constructed an input vector, $\mathbf{F}_{\text{layer}}$, which contains the following layer-specific characteristics:

$$\mathbf{F}_{\text{layer}} = \{f_1, f_2, \ldots, f_n\} \quad (1)$$

where each $f_i$ represents an individual feature relevant to the computational characteristics of a specific layer type, as shown below:

- **Dense Layer:** Input and output dimensions and batch size.

- **Convolution Layers:** Kernel size, padding, stride, input and output channels and spatial dimensions.
- **Recurrent Layers:** Sequence length, hidden state size and the number of layers recorded.
- **Layer Normalization:** Input dimension and batch size.
- **Embedding Layer:** Vocabulary size and embedding dimension.
- **Multi-Head Attention:** Number of Attention heads, query and key dimensions, output dimension and sequence length.

## 3.2 PreNeT FEATURES

In addition to layer-specific characteristics, computational features are also critical to accurately predicting training times. These features define each layer's *arithmetic intensity* and *computational demands*, providing a more comprehensive estimate of the time it will take to execute them. In the following, we will further expand on the main computational features, which are the layer's arithmetic intensity and computation demands.

*3.2.1 CO: Computational Operations.* This measures the number of arithmetic operations each layer performs and reflects the computational complexity involved. For each type of layer, we computed a computational operation based on the various configuration parameters. These calculations provide a reasonable estimate of the computational demand for each specific layer; however, they can vary in certain situations. This variation can be attributed to several factors, such as branching operations that depend on the size of the data input, as well as differences in learning frameworks like PyTorch and TensorFlow. Additionally, low-level hardware optimizations can affect the actual computational operations during execution. Furthermore, hardware-specific elements, such as parallel processing and caching, can also influence the calculations related to computational operations.

- **Dense Layer:** The calculation of computational operations for a fully connected layer is derived from the product of input dimension, output dimension, and batch size. For a dense layer with input dimension $d_{in}$, output dimension $d_{out}$, and batch size $b$, the computational operations can be calculated as:

$$CO_{Dense} = b \times d_{in} \times d_{out} \quad (2)$$

- **Convolutional Layer [20]:** Computational Operations for convolutional layers involve the kernel size, spatial dimensions and channel sizes as stated in [35]. For a convolutional layer with batch size $b$, kernel size $k \times k$, input channels $c_{in}$, output channels $c_{out}$, and spatial dimensions (height $h$, width $w$), the computational operations can be calculated as:

$$CO_{Conv} = b \times h_{out} \times w_{out} \times k^2 \times c_{in} \times c_{out} \quad (3)$$

where $h_{out}$ and $w_{out}$ are the output dimensions.
- **Recurrent Layers:** Since RNN (Recurrent Neural Network), LSTM (Long Short-Term Memory) [6] and GRU (Gated Recurrent Unit) [4] have distinct internal structures and gating mechanisms, the computational operations are calculated in three separate formulations:

  – **RNN (Recurrent Neural Network):** For a RNN layer with sequence length $s$, hidden size $h$, input size $d$, and batch size $b$:

  $$CO_{RNN} = 2 \times b \times s \times h \times (h + d) \quad (4)$$

  – **LSTM (Long Short-Term Memory):** For a LSTM layer with the same parameters but including four gates (input, input modulation, forget, and output):

  $$CO_{LSTM} = 4 \times b \times s \times h \times (h + d) \quad (5)$$

  – **GRU (Gated Recurrent Unit):** For a GRU layer, which includes three gates (reset, update, and forget):

  $$CO_{GRU} = 3 \times b \times s \times h \times (h + d) \quad (6)$$

- **Multi-Head Attention Layer [44]:** The computational complexity of the Multi-Head Attention Layer, which is a core component of the Transformer and LLM architectures, depends on several critical parameters, including the number of Attention heads $h$, the sequence length $s$, and the dimension $d$ of the query, key, and value vectors. Each head processes the sequence independently, and the outputs are then concatenated to form the final Attention representation. For a Multi-Head Attention layer, the computational operation calculation is given by:

$$CO_{Attention} = h \times s^2 \times d \quad (7)$$

- **Embedding Layer [22]:** Since there is no floating-point multiplication or additions performed in the Embedding layer, we have developed a practical computational feature for this layer. Given a batch size $B$ and Sequence Length $S$, each token in the input sequence is mapped to an embedding vector of size $D_{emb}$. The Computational Feature for the Embedding layer is calculated as:

$$CO_{Embedding} = B \times S \times D_{emb} \quad (8)$$

*3.2.2 CT: Computational Time.* After the calculation of the computational operations for each layer, inspired by [35], we introduce **Computational Time** as the additional feature for enhancing the training time prediction accuracies. Computational Time represents the time it takes for certain hardware (bounded by GPUs) to process the total operations, which varies significantly throughout various hardware setups. For example, Nvidia's Tesla P4 GPU [24] carries a floating-point processing power of 5.704 TFLOPs, while Nvidia's RTX4090 GPU [28] has the power of 82.59 TFLOPs. By integrating both computational operations and time, our model outperforms state-of-the-art approaches for training time predictions by learning each layer's computational requirements within the context of hardware performance in more detail. As mentioned in 3.2.1, $CO_{(L)}$ represents the operations for a specific layer with capturing computational demands in various architecture settings. $G_{FLOPs}(D)$ denote the GPU's theoretical peak FLOPs rate for hardware $D$. The $G_{FLOPs}$ values for each GPU is shown in the last column of Table 2.

Then, Computational Time $CT(L, D)$ for layer $L$ and hardware $D$ is given by:

$$CT(L, D) = \frac{CO_{(L)}}{G_{FLOPs}(D)} \quad (9)$$

PreNeT: Leveraging Computational Features to Predict Deep Neural Network Training Time

Table 2: Hardware Specifications

| Name | Provisioning | Boost Clock (MHz) | Memory (GB) | Memory Type | Bus | Cores | GFLOPs |
|---|---|---|---|---|---|---|---|
| Nvidia P4 [24] | GCP | 1114 | 8 | GDDR5 | PCIe 3.0 | 2560 | 5700 |
| Nvidia P100 [23] | GCP | 1329 | 16 | HBM2 | PCIe 3.0 | 3584 | 9526 |
| Nvidia RTXA4000 [27] | Local | 1560 | 16 | GDDR6 | PCIe 4.0 | 6144 | 19170 |
| Nvidia T4 [26] | GCP | 1590 | 16 | GDDR6 | PCIe 3.0 | 2560 | 8141 |
| Nvidia V100 [25] | GCP | 1380 | 16 | HBM2 | PCIe 3.0 | 5120 | 14130 |
| Nvidia L4 [29] | GCP | 2040 | 24 | GDDR6 | PCIe 4.0 | 7424 | 30290 |
| Nvidia RTX4090 [28] | Local | 2520 | 24 | GDDR6X | PCIe 4.0 | 16384 | 82580 |

By treating Computational Time as an additional feature and combining it with the computational operation calculations per layer, we provide a richer set of input features that enhances the model's accuracy in estimating the training time.

### 3.3 CM: COMPUTATIONAL MEMORY

In addition to the layer-specific and computational features, it is important to consider memory usage, which affects the training time of machine learning models. For example, the memory usage of models like GPT-3[2], which has 175 billion parameters that require over 700GB of memory, has a direct impact on their training time, where higher memory utilization results in limitations of batch size, which reduces parallelism. Therefore, memory optimization is a key factor in achieving efficient execution times. For each layer type, we have captured the memory required for storing the layer's inputs and output tensors and weights and provide a detailed view of the layer's memory footprint.

For each layer type, the following set of memory-related features are computed:

- **Memory for Weights:** For layers such as dense and convolutional layers, the memory required to store the weights is calculated as:

$$\text{Memory}_{\text{weights}} = D_{in} \times D_{out} \quad (10)$$

where $D_{in}$ is the input dimension and $D_{out}$ is the output dimension.

- **Memory for Input:** The memory required to store the input tensor, based on the batch size $B$ and the input dimension $D_{in}$, is calculated as:

$$\text{Memory}_{\text{input}} = B \times D_{in} \quad (11)$$

- **Memory for Output:** The memory required to store the output tensor, based on the batch size $B$ and the output dimension $D_{out}$, is calculated as:

$$\text{Memory}_{\text{output}} = B \times D_{out} \quad (12)$$

These memory-related features are important for capturing the memory footprint for each specific layer during the training process, especially in larger models where the memory limitations can significantly impact the training time.

### 3.4 AGGREGATED TRAINING TIME PREDICTION

To accurately estimate the total training time for a complete model, we have used a layer-wise and batch-wise summation strategy. In the final step, we aggregate the predicted runtime for each layer across all batches within an epoch, ensuring that individual layer complexities and batch configurations are accounted for. By capturing this level of granularity, our model confidently delivers robust predictions for a wide range of architectures and hardware setups.

Let:

- $L$: Total number of layers in the model
- $B$: Total number of batches in one epoch
- $T_{\text{pred},i}(j)$: Predicted time for layer $i$ on batch $j$

The total predicted runtime of an epoch, denoted as $T_{\text{epoch}}$, is then calculated as:

$$T_{\text{epoch}} = \sum_{j=1}^{B} \sum_{i=1}^{L} T_{\text{pred},i}(j) \quad (13)$$

This cumulative calculation method enables our model to utilize the features of individual layers, along with memory and computational requirements, to estimate the training times accurately. By adding both layer-wise and batch-wise times, this approach adapts effectively to various neural network architectures, providing precise predictions across different GPU configurations.

## 4 EVALUATION AND DISCUSSION

We evaluate the predictive accuracy of our model, PreNeT, by analyzing its performance across various architectures and hardware configurations. We performed an extensive set of experiments to assess how well PreNeT generalizes to different neural network layers, computational complexities, and hardware configurations. By comparing the predicted training times against the actual completion time, we have assessed PreNeT's predictive performance and its capacity to handle unseen configurations in terms of GPUs and layer input parameters. To ensure reproducibility, we have provided the code and data that we used for our experimental evaluation in an anonymous Github repository.[1]

### 4.1 EXPERIMENT SETUP

As shown in Table 2, we have run our experiments using both Google Cloud Platform's (GCP) GPUs within the same time zone and our locally accessible RTX GPUs. By running the experiments across both environments, it allows us to assess the generalization of our predictions. Our benchmarking setup covers various GPU configurations with variations in core count, memory capacity types and clock speeds. With our selection of GPUs from different categories, we ensure that PreNeT captures the distinction in computational capacity and memory handling across various types of hardware settings. Due to the features we feed into our model, it is crucial to use generated data from a wide range of GPUs. For instance, GPUs with higher core counts and faster memories can handle complex models more efficiently than others.

---
[1] https://github.com/pacslab/PreNeT

Table 3: Input Parameters for Each Layer Type

| Layer Type | Parameter | Range | Models Using These Parameters |
|---|---|---|---|
| Dense | Batch Size ($B$) | 1 to 64 | BERT[8], ResNet[15] |
| | Input Dimension ($D_{in}$) | 1 to 4096 | GPT-3[2], BERT |
| | Output Dimension ($D_{out}$) | 1 to 4096 | AlexNet[19], VGG[39] |
| Convolutional | Batch Size ($B$) | 1 to 64 | ResNet, VGG |
| | Input Dimension (Matrix Size) | 1x1 to 512x512 | AlexNet, EfficientNet[43] |
| | Kernel Size ($K$) | 1x1 to 7x7 | ResNet, Inception[42] |
| | Stride | 1, 2 | MobileNet[16], ResNet |
| | Input Padding | 0, 1 | AlexNet, VGG |
| Attention | Batch Size ($B$) | 1 to 64 | Transformer[44], BERT |
| | Sequence Length ($L$) | 64 to 512 | GPT-3, BERT |
| | Embedding Dimension ($D_{emb}$) | 128 to 1024 | T5[36], GPT-2 |
| | Number of Heads ($H$) | 1 to 16 | Transformer, BERT |
| Embedding | Batch Size ($B$) | 1 to 64 | Word2Vec[22], BERT |
| | Vocabulary Size | 10,000 to 500,000 | FastText[1], GPT-2 |
| | Embedding Dimension ($D_{emb}$) | 64 to 1024 | Word2Vec, BERT |
| | Sequence Length ($L$) | 64 to 512 | GPT-3, T5 |
| Recurrent | Batch Size ($B$) | 1 to 64 | LSTM[6], GRU[4] |
| | Sequence Length ($L$) | 1 to 512 | Seq2Seq[41], Transformer |
| | Hidden Dimension ($H$) | 128 to 1024 | BiLSTM[38], RNN-T[13] |
| | Bidirectional | True, False | BiLSTM |
| Layer Normalization | Batch Size ($B$) | 1 to 64 | Transformers, GPT |
| | Input Dimension ($D_{in}$) | 64 to 1024 | BERT, T5 |

## 4.2 PARAMETER SELECTION ACROSS LAYER TYPES

For each layer type, we have systematically generated a wide range of input parameters to ensure comprehensive coverage of the possible configurations. These parameters include elements such as batch size and feature-specific parameters (e.g., the number of heads for the Attention layer). The ranges of these parameters were selected using the inputs of various machine learning models such as GPT-3, BERT, VGG[39] as shown in Table 3. To achieve an accurate estimate for each layer type, we ran a random combination of the configurations of over 20, 000 iterations with five repetitions in each GPU to calculate the mean runtime. To further reduce noise and variability in the measurements, our data collection was conducted at different hours of the day. The collected data was split into a training dataset (80%), a validation data set (10%) and a test data set (10%) for calculating the RMS values for each layer type using various machine learning techniques and comparing our results with the baseline in 5.1, followed by testing our best PreNeT performing model with unseen GPU configurations and how it behaves in those settings as discussed in 5.2.

## 4.3 EVALUATION METRIC

In our work, we utilize the **Root Mean Square Error (RMSE)** as the primary evaluation metrics in order to measure the accuracy of PreNeT training time prediction as below:

$$\text{RMSE} = \sqrt{\frac{1}{N} \sum_{i=1}^{N} (\hat{y}_i - y_i)^2} \quad (14)$$

where $N$ represents the number of the data samples, $\hat{y}_i$ is the predicted training time for the $i$-th sample, and $y_i$ is the actual observed training time from PreNeT. RMSE provides a reliable measurement of prediction accuracy, as it penalizes larger errors more heavily, reflecting the impact of significant deviations on model performance. Lower RMSE values indicate higher prediction accuracy. This metric is essential for validating our model's generalization ability across different architectures and hardware configurations.

## 4.4 CF-PRED MACHINE LEARNING MODELS

To measure the difference in prediction accuracies for the set of features that we have defined in our work, we have trained various machine learning models to see which one is giving the best performance for each layer type. Overall, random forest and fully connected layer-based models performed the best for predicting the training time per epoch for the defined layer types as discussed in 5.1. In our work, we have compared and observed the results using Linear Regression, Random Forest, Gradient Boosted Decision Trees (GDBT), eXtreme Gradient Boosting (XGBoost) and Fully Connected Layer (MLP). The MLP model consists of six hidden layers, with the following number of neurons in each layer: 32, 64, 128, 128, 128, and 128 with ReLU activation functions. To prevent overfitting, a dropout layer with a rate of 0.2 is included right before the output layer. The model is trained for 300 epochs with a batch size of 128 and a learning rate of 0.001.

Table 4: Performance Comparison Across Different Layers and Models (RMSE) | CO: Computational Operations
COCM: Computational Operations and Memory | COCMCT: Computational Operations, Memory and Time

| | Attention Layer | | | | | Convolution Layer | | | | |
|---|---|---|---|---|---|---|---|---|---|---|
| Architecture | Baseline | CO | COCM | COCMCT | Improvement % | Baseline | CO | COCM | COCMCT | Improvement % |
| **Linear Regression** | 4.943 | 4.772 | 4.524 | 3.005 | 39.21% | 18.612 | 17.650 | 17.186 | 14.094 | 24.27% |
| **Random Forest** | 0.543 | 0.483 | 0.353 | 0.382 | 29.65% | 8.328 | 8.159 | 8.220 | **7.909** | 5.03% |
| **GDBT** | 2.588 | 1.750 | 1.015 | 0.709 | 72.60% | 13.902 | 11.17 | 10.47 | 9.377 | 32.55% |
| **XGBoost** | 0.56 | 0.461 | 0.320 | 0.329 | 41.25% | 8.935 | 8.405 | 8.168 | 8.110 | 9.23% |
| **Fully Connected Layer** | 0.335 | 0.319 | 0.318 | **0.312** | 6.87% | 9.633 | 10.324 | 9.375 | 9.093 | 5.61% |
| | Embedding Layer | | | | | Fully Connected Layer | | | | |
| **Linear Regression** | 3.984 | 4.079 | 3.787 | 3.734 | 6.28% | 0.841 | 0.817 | 0.796 | 0.772 | 8.21% |
| **Random Forest** | 0.583 | 0.425 | 0.378 | 0.382 | 34.48% | 0.127 | 0.130 | 0.126 | **0.117** | 7.87% |
| **GDBT** | 1.943 | 1.878 | 1.074 | 1.066 | 45.14% | 0.383 | 0.378 | 0.244 | 0.240 | 37.34% |
| **XGBoost** | 0.576 | 0.536 | 0.383 | 0.387 | 32.81% | 0.136 | 0.145 | 0.130 | 0.125 | 8.09% |
| **Fully Connected Layer** | 0.363 | 0.389 | 0.274 | **0.284** | 21.76% | 0.132 | 0.130 | 0.136 | 0.129 | 2.27% |

Table 5: Performance Comparison for RNN

| RNN Layer | | | | | |
|---|---|---|---|---|---|
| Architecture | Baseline | CO | COCM | COCMCT | Improvement % |
| **Linear Regression** | 1.204 | 1.053 | 1.028 | 0.853 | 29.15% |
| **Random Forest** | 0.387 | 0.376 | 0.269 | 0.368 | 30.49% |
| **GDBT** | 0.905 | 0.515 | 0.505 | 0.453 | 49.94% |
| **XGBoost** | 0.351 | 0.291 | 0.269 | 0.28 | 23.36% |
| **Fully Connected Layer** | 0.362 | 0.341 | 0.262 | **0.219** | 39.50% |

## 5 RESULTS

In the following section, we present the results achieved from our framework. The first set of results is the RMSE value achieved from predicting the training time per epoch for each layer type, followed by a description of how our framework handles unseen hardware configurations and how it enhances the prediction accuracy for VGG-16 [39] and BERT models.

### 5.1 COMPUTATIONAL FEATURES IMPACT

We compared our results to Justu's work [18], which employs a similar strategy for predicting ML model training times for each epoch. In their work, CNN, Dense, and RNN layer types have been investigated. By introducing our computational features, we have improved the accuracy of predicting the training time per epoch, as mentioned in the following tables. Our baseline did not investigate Attention, Embedding and Normalization layers, which adds an important contribution to our work by benchmarking and introducing the computational features of these layers. Using the benchmarking data for all layer types, we first applied the traditional approach of applying various machine learning techniques to predict the training times (similar to [18]) and then fed our features to these models and compared the improvements as shown in Table 4 and Table 5.

Based on the observations in these tables, PreNeT's approach in predicting the training time per layer type significantly improves the errors compared to our baseline up to **72.60%**. In certain cases, as highlighted Table 4 and Table 5, such as Attention, RNN and Embedding Layers, the combination of using the computational time along with computational memory and operations does not improve the prediction accuracies compared to not using the computational time feature along with the other computational features. This is due to the nature of these layers, which rely heavily on memory bandwidth and access patterns rather than raw computational throughput. For instance, the complicated matrix operations that are required in the Multi-Head Self-Attention results in more bottlenecks in memory computations compared to operations.

Similarly, Embedding layers are relatively lightweight in terms of computation, as their primary focus involves lookups in the Embedding tables. Having said that, these lookups are memory-bound operations and not computational-bound, which results in a much greater impact on training times. Also, this applies to RNNs (especially LSTM and GRU) because they maintain hidden states across multiple time steps, which requires extensive memory resources. As can be observed, the combination of these features has a much higher impact on CNN and Fully Connected Layers, where the compute-to-memory ratio is higher than the Attention and Embedding Layers. In CNN layers, the main operations involve convolutional kernels sliding over the input data, which allows for extensive parallel processing on GPUs with optimized cores like CUDA or Tensor cores. This structure means that CNNs can leverage the hardware's maximum computational capacity, which results in making the computational time feature a dominant element for training time prediction models. In addition, fully connected layers, due to the high density of operations, such as matrix multiplications across the input and output neurons, benefit from the computational time feature.

In our work, we have trained various machine learning models such as Linear Regression, Random Forest, GDBT, XGBoost, and Fully Connected Layer to compare which models perform the best for each layer type training time prediction. Our results show that fully connected neural networks for Attention with RMSE of $0.312\,ms$, Embedding with RMSE of $0.284\,ms$ and RNN-based layers with RMSE of $0.262\,ms$ perform the best due to the complex, non-linear dependencies in these layers. However, for CNN with RMSE of $7.909\,ms$ and Fully Connected Layers with RMSE of $0.117\,ms$, random forests yield better training time predictions. This is due to their structured operations, which allow random forests to capture well by creating decision trees based on specific thresholds in the feature space.

## 5.2 PERFORMANCE ON UNSEEN HARDWARE CONFIGURATIONS

For our evaluation, we tested PreNeT performance on unseen hardware configurations. Specifically, we trained PreNeT using the benchmarking data and computational features from five GPUs: Nvidia V100, P100, P4, T4, and RTX 4090 and then evaluated its training time predictions on two unseen GPUs: Nvidia L4 and RTX A4000. It is important to note that our evaluation included GPUs from both local and cloud setups, allowing us to observe how PreNeT adapts to different provisioning setups.

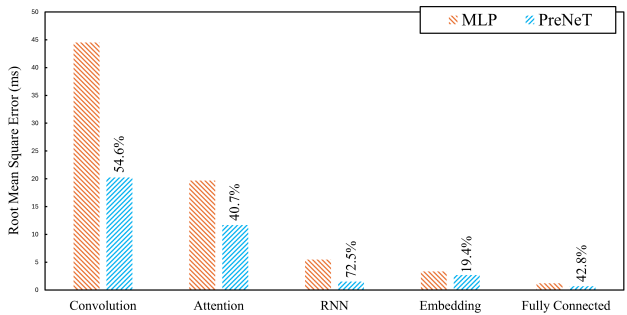

**Figure 2: PreNeT vs. Baseline MLP on Unseen Hardware**

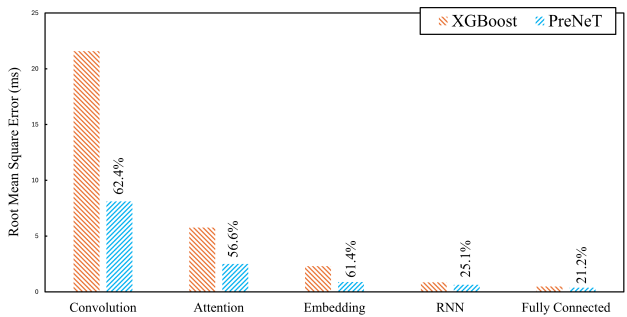

**Figure 3: PreNeT vs. Baseline XGBoost on Unseen Hardware**

By introducing the computational features for predicting the training time across various layers, as illustrated in Figure 2 and Figure 3, PreNeT outperforms the baseline approach. The improvement percentages by using PreNeT features are shown above the bars for each layer type, which ranges from **19.4%** for the Embedding layers at lowest up to **72.5%** for the RNN layers. In these trials, in the first case, an MLP-based PreNeT model was trained using all the computational features, and similarly, in the second case, an XGBoost-based PreNeT model was trained on the same set of features. The RMSE clearly shows how promising PreNeT is in predicting the training time of various architectures, even for unseen GPUs. We have also achieved similar improvements using other machine learning techniques, but the most notable improvements were made using these two machine learning models. Since CNN layers rely heavily on the optimized memory access patterns and efficient utilization of CUDA or Tensor cores and the way different GPUs handle these optimizations, we still get an RMSE of $8.114 ms$.

## 5.3 FULL MODEL PREDICTION

We have chosen VGG-16 and BERT[8] models for testing PreNeT training time predictions. For VGG-16, our model was trained using a dataset size of 1024. Each image in this dataset was randomly created to match the size of the images in ImageNet [7]. Similarly, we used a dataset size of 1024 for BERT and trained our model for 100 epochs. Building on the work of [18], we first estimate the training time for forward and backward passes through each layer of both architectures. We then combine these results to compare the predicted training time per epoch with the actual training time. We have chosen the best model for each layer type with our defined computational features. To report the improvement of PreNeT predictions compared to the baseline, as illustrated in Figure 4 to Figure 6, we have reported how PreNeT is outperforming Justu's work for training the VGG-16 model. For example, using Justu's method, the prediction error for training VGG-16 was highest at 82.03% with a batch size of 64 using P100, while our method achieved an error of 27.65%. Also, we achieved 22.81% to 38.42% improvements for predicting the training time of BERT with RMSEs ranging from $3.02s$ to $2.7s$ with batch sizes 16 to 64 by using Nvidia RTX4090 GPU. As mentioned earlier, our baseline did not consider the features of transformer-based model layer types, but we have applied their strategy for these layers and still outperformed them using our computational features. We have trained the best-performing machine learning model (Random Forest) for 100 epochs using our computational features for predicting the training time of VGG-16 in each epoch and reported the results and improvements using the input batches of 16 (Figure 4), 32 (Figure 5) and 64 (Figure 6). The observed results clearly show the impact of our defined computational features on the training time predictions. For instance, our prediction errors at its lowest are at **4.50%** using Nvidia's V100 for training VGG-16 with its' complex architecture. As observed, our RMSE values are at their highest for CNN layers, and we have tested how our framework predicts the training time of VGG-16, which consists of 16 layers, including 13 convolutional and 3 fully connected layers. However, even for the architectures that carry these types of complexities, the reported results show how promising our approach is, which also outperforms the baseline in all batch sizes, as it could be seen.

## 6 RELATED WORK

In recent years, predicting training times and computational costs of deep learning models has been an active area of research. Several studies have attempted different coping methods for that challenge [10, 11, 18]. In one of the early works in this domain [18], the authors presented a data-based model of how long it takes the key layers in deep neural networks (convolution, fully connected, etc.) to execute. They demonstrated that their method can accurately predict the time needed for training a single batch, allowing for extrapolation over an entire epoch. Nevertheless, their work normalized deep learning architecture, and they didn't consider the distinguishing features of transformer-based models such as LLMs.

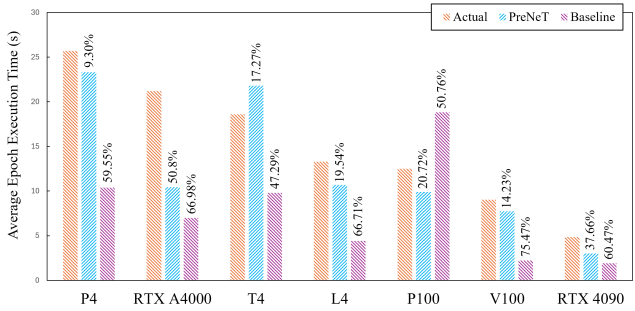

**Figure 4: Prediction Errors for PreNeT and Baseline vs. Actual Training Times (Batch Size 16)**

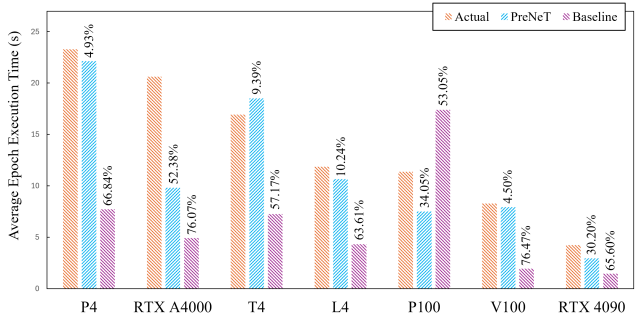

**Figure 5: Prediction Errors for PreNeT and Baseline vs. Actual Training Times (Batch Size 32)**

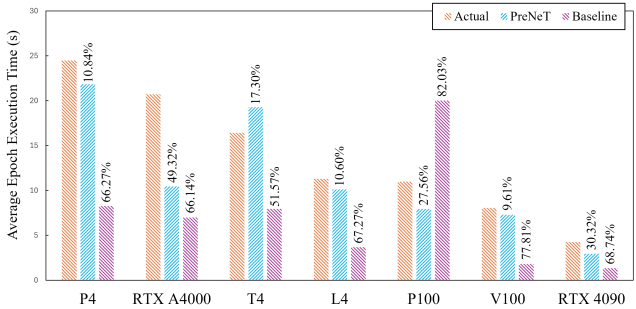

**Figure 6: Prediction Errors for PreNeT and Baseline vs. Actual Training Times (Batch Size 64)**

Habitat [11] utilizes the repetitive nature of DNN training curves and runtime information from GPU hardware configurations of different prediction performances. This work yielded encouraging results, but it was still limited to conventional DNN types and had little to say about the challenges faced by large-scale transformer models. With the aid of graph neural networks, DNNPerf [10] formulates a deep learning model as a directed acyclic computation graph. In addition, DNNPerf incorporates many performance-related features, accommodating both the vertex-level and graph-level aspects. While DNNPerf displayed excellent accuracy in predicting the training times, it was evaluated primarily on traditional DNN architectures and didn't target transformer-based models specifically.

Zancato et al. [47] addressed the challenge of determining the number of optimization steps necessary for a pre-trained deep network to converge to a specific loss level, particularly within the context of transformer-based models. They utilized the fact that the training dynamics of a deep network during fine-tuning can be approximately modelled by a linearized model. By doing this, they could predict the time required for training without actually carrying it out, which can mean substantial cost savings. The main problem with their work is that they did not provide any sort of in-depth analysis as to how different hardware setups and architectural elements will affect system performance.

Our study aims to address this gap by creating a framework that collects detailed training time information from various GPU models to effectively train different machine learning algorithms to predict training durations precisely with the addition of computational features for a wide range of layer types. Our work is inspired by [18] where the training time is predicted based on two factors: The amount of time needed to do a single epoch — forward and backward passes through your data together and the number of epochs you will need to reach the required accuracy level. Although both are important, our work specifically targets predicting a single epoch's training time for deep learning models, including Transformers. Focusing on this informs us how various model architectures with different hardware configurations and hyper-parameter settings will influence the training time.

To the best of our knowledge, no other approach exists for predicting the training time of machine learning models (including transformer-based models such as LLMs) in an atomic way by incorporating the computational complexity of various types of layers. It is important to note that we have investigated the impact of all types of layers used in the architecture of various ML models and have efficiently predicted the training time per epoch. Our strategy can also be applied to new architectures due to its design.

## 7 CONCLUSION & FUTURE WORK

In this paper, we presented PreNeT, a novel predictive framework that addresses the critical challenge of optimizing deep learning model training, particularly emphasizing Transformer-based architectures. Through comprehensive empirical evaluation, we have demonstrated PreNeT's superior capability in predicting training requirements and duration across diverse hardware configurations. The framework's distinctive approach to integrating computational metrics, encompassing arithmetic operations, memory utilization, and layer-specific parameters, has proven instrumental in achieving a 72% improvement in prediction accuracy over existing state-of-the-art solutions.

The significance of PreNeT extends beyond mere performance metrics. By enabling accurate predictions for previously unexamined hardware infrastructures, PreNeT provides researchers and practitioners with a powerful what-if analysis tool for proactive resource allocation and cost optimization. This capability is particularly valuable given the increasingly complex landscape of deep learning infrastructure and the substantial computational demands of modern neural architectures.

Looking ahead, PreNeT establishes a robust foundation for future research in training optimization and resource prediction for deep learning systems. More specifically, at this stage, PreNeT only optimizes the training at the epoch level. It does not predict the number of epochs needed to achieve certain accuracy. In our future work, we plan to chain PreNeT with another framework to model and optimize the total training time for large and deep neural networks.